%% file: Main.tex
\documentclass{ecai}
\pdfoutput=1

\usepackage[utf8]{inputenc} 
\usepackage[T1]{fontenc}    
\usepackage{hyperref}       
\usepackage{url}            
\usepackage{booktabs}       
\usepackage{amsfonts}       
\usepackage{nicefrac}       
\usepackage{microtype}      
\usepackage{graphicx}
\usepackage{comment}
\usepackage{bm}
\usepackage{amsmath}
\usepackage{amssymb}
\usepackage{caption}
\usepackage{adjustbox}
\usepackage{multirow}
\usepackage{xcolor}
\usepackage{array}
\usepackage{subfig}

\usepackage{array}
\newcolumntype{P}[1]{>{\centering\arraybackslash}p{#1}}
\newcolumntype{M}[1]{>{\centering\arraybackslash}m{#1}}

\DeclareMathOperator*{\argmax}{argmax}
\DeclareMathOperator*{\argmin}{argmin}

\newtheorem{mydef}{Definition}

\begin{document}

\title{Toward Metrics for Differentiating Out-of-Distribution Sets}

\author{
  Mahdieh Abbasi\institute{Universit\'e Laval, Canada,  \texttt{mahdieh.abbasi.1@ulaval.ca}},  Changjian Shui\institute{Universit\'e Laval, Canada,  \texttt{changjian.shui.1@ulaval.ca}},  Arezoo Rajabi\institute{Oregon State University, USA, \texttt{rajabia@oregonstate.edu}}, Christian Gagn\'e\institute{Canada CIFAR AI Chair / Universit\'e Laval, Canada,\protect\\ \texttt{christian.gagne@gel.ulaval.ca}}~ \textrm{and} Rakesh B. Bobba\institute{Oregon State University, USA , \texttt{rakesh.bobba@oregonstate.edu}}
}

\maketitle

\begin{abstract}
Vanilla CNNs, as uncalibrated classifiers, suffer from classifying out-of-distribution (OOD) samples nearly as confidently as in-distribution samples. To tackle this challenge, some recent works have demonstrated the gains of leveraging available OOD sets for training end-to-end calibrated CNNs. However, a critical question remains unanswered in these works: how to differentiate OOD sets for selecting the most effective one(s) that induce training such CNNs with high detection rates on unseen OOD sets? To address this pivotal question, we provide a criterion based on generalization errors of Augmented-CNN, a vanilla CNN with an added extra class employed for rejection, on in-distribution and unseen OOD sets. However, selecting the most effective OOD set by directly optimizing this criterion incurs a huge computational cost. Instead, we propose three novel computationally-efficient metrics for differentiating between OOD sets according to their ``protection'' level of in-distribution sub-manifolds. We empirically verify that the most protective OOD sets -- selected according to our metrics -- lead to A-CNNs with significantly lower generalization errors than the A-CNNs trained on the least protective ones. We also empirically show the effectiveness of a protective OOD set for training well-generalized confidence-calibrated vanilla CNNs. These results confirm that 1) all OOD sets are not equally effective for training well-performing end-to-end models (i.e., A-CNNs and calibrated CNNs) for OOD detection tasks and 2) the protection level of OOD sets is a viable factor for recognizing the most effective one. Finally, across the image classification tasks, we exhibit A-CNN trained on the most protective OOD set can also detect black-box FGS adversarial examples as  their distance (measured by our metrics) is becoming larger from the protected sub-manifolds.
\end{abstract}

\input{Intro}

\input{proposed_method}
\input{Evaluation}

\input{RelatedWork}

\input{conclusion}

{\small
\bibliographystyle{ecai}

\bibliography{Reference}
}


\end{document}

%% file: Intro.tex
\section{Introduction}
\label{sec:intro}





In supervised learning, it is generally assumed that a training set and a held-out test set are drawn independently from the same data distribution, called \emph{in-distribution set}. While this assumption can be true for controlled laboratory environments, it rarely holds for many real-world applications, where the samples can be drawn from both in-distribution and from other distributions, called \emph{out-of-distribution} (OOD) data, which contains samples that are semantically and statistically different from those in-distribution. In the presence of OOD samples, it is important to have a model able to distinguish them in order to make reliable decisions. However, it has been shown that state-of-the-art (vanilla) deep neural networks (e.g., CNN) are uncalibrated such that they are making predictions for OOD samples with a confidence that is as high as those of in-distribution samples, making them indistinguishable from each other~\cite{guo2017calibration,hein2019relu}. For safety-critical real-world applications such as self-driving cars, using vanilla CNNs that tend to confidently make wrong decisions for such unknown OOD samples can lead to serious safety and security consequences.

To tackle this challenge, post-processing approaches~\cite{devries2018learning,hendrycks2016baseline,lee2018simple,liang2017principled} attempt to transform the confidence of predictions made by pre-trained vanilla CNNs in order to create a gap between the confidence for OOD samples and that of in-distribution ones. Despite their simplicity and efficiency, their performances depend on several additional hyper-parameters such as temperature, magnitude of additive noise, or the parameters of an auxiliary regression function, which should be carefully tuned for each OOD set.

More recently, some researchers~\cite{Meinke2020ood,bevandic2018discriminative,hendrycks2018deep,lee2017training,masana2018metric} have proposed \emph{end-to-end} calibrated CNNs-based models for OOD detection. For instance, calibrated vanilla CNNs~\cite{hendrycks2018deep,hein2019relu,lee2018simple} are trained to make uncertain predictions for OOD samples while still confidently and correctly classifying in-distribution ones. To train these models, some authors~\cite{Meinke2020ood,bevandic2018discriminative,hendrycks2018deep,masana2018metric} have leveraged a naturalistic OOD set\footnote{We simply drop naturalistic and instead call it OOD set throughout the paper.}, which seemingly is selected \emph{manually} from among many available ones, without providing a systematical justification for their selection. Thus, the following question remains unaddressed in these works: \emph{how to differentiate among OOD sets w.r.t.\ a given in-distribution task with the goal of selecting the most proper one, which in turn induces a well-generalized calibrated model with high detection rate of \emph{unseen} OOD sets?}. 

Besides the confidence-calibrated vanilla CNN~\cite{Meinke2020ood,hendrycks2018deep,lee2017training} as end-to-end model for OOD detection task, the classical idea of adding an explicit rejection class~\cite{abbasi2018towards,bendale2016towards,da2014learning,gunther2017toward} is also an interesting end-to-end approach. Indeed, such augmented classifiers can directly reject OOD samples by classifying them to the extra class, while correctly classifying in-distribution samples. In addition to the calibrated vanilla CNN, we exploit A-CNN as an end-to-end model for OOD detection task.

However, without having a principle for selecting the right OOD sets among those available, training \emph{well-generalized A-CNN and calibrated vanilla CNN} is challenging. Since using a randomly selected OOD set does not necessarily lead to a model with a high detection rate of unseen OOD sets (i.e., generalization ability) as we later show in our experiments. It has also been shown in~\cite{lee2017training}, where using SVHN as OOD set for CIFAR-10 is leading to an A-CNN with inferior generalization properties. Moreover, simply using a union of an enormous number of OOD sets not only creates an extremely unbalanced dataset, but also makes training of these models computationally infeasible. 

Although Hendrycks et al.~\cite{hendrycks2018deep} have conjectured diversity for characterizing a proper OOD set, in this paper, our main focus is to answer concretely the aforementioned question, i.e., how to differentiate between OOD sets in order to select a proper one. At first, we provide a formal criterion in the form of generalization errors of A-CNN for differentiating OOD sets and selecting the most effective one. Using this, an OOD set is recognized as a proper (effective) if it leads to training of A-CNN with low generalization errors for both in-distribution and unseen OOD sets. However, selecting a proper OOD set by directly optimizing this selection criteria is computationally very expensive due to the existence of tremendous number of OOD sets. To overcome this, we propose some metrics that can be efficiently computed using a pre-trained vanilla CNN. We drive our metrics according to the following intuition: a proper (effective) OOD set should cover sub-manifolds of an in-distribution task, which can be achieved by the penultimate layer of a vanilla CNN trained on it. Thus, we design our metrics to measure the degree of protectiveness of sub-manifolds by OOD sets for selecting the most protective one. Indeed, protecting in-distribution sub-manifolds by OOD samples allows for rejecting \emph{automatically} the unseen OOD sets which are located relatively far away from the protected in-distribution sub-manifolds, as shown in Figure \ref{toyexample}. Therefore, the protection level of OOD sets can be a viable factor for differentiating and selecting of OOD sets with the aim of obtaining a well-generalized A-CNN and calibrated vanilla CNN.

Our contributions in this paper can be outlined as follows:
\begin{itemize}

\item We provide a formal definition (with the use of A-CNN) for precisely differentiating OOD sets according to their induced generalization errors on unseen OOD sets and in-distribution set.

\item We are first to propose novel quantitative metrics for differentiating OOD sets with the aim of selecting the most protective OOD set w.r.t.\ a given in-distribution set. These metrics, namely \textbf{Softmax-based Entropy (SE)}, \textbf{Coverage Ratio (CR)} and \textbf{Coverage Distance (CD)}, can be \emph{efficiently} computed using a vanilla CNN trained on the given in-distribution task.
    
\item In an extensive series of experiments on image and audio classification tasks, we empirically show that A-CNNs and calibrated vanilla CNNs trained on the most protective OOD set have higher detection rates (lower generalization error) on unseen OOD sets  in comparison with those trained on the least protective OOD set.

\item We exhibit that A-CNN trained on the most protective OOD set (i.e., A-CNN$^{\star}$) can also detect black-box FGS adversarial examples generated by a relatively large magnitude of noise, while vanilla CNN and the A-CNN trained on the least protective OOD set are still incorrectly classifying them. We show this occurs as the distance (measured by CD) of FGS adversaries is increased from the protected sub-manifolds.

\end{itemize}

%% file: proposed_method.tex
\begin{figure*}[ht]
   \centering
    \subfloat{\includegraphics[width=0.2\textwidth,trim= 1cm 0cm 0cm 1cm, clip=true]{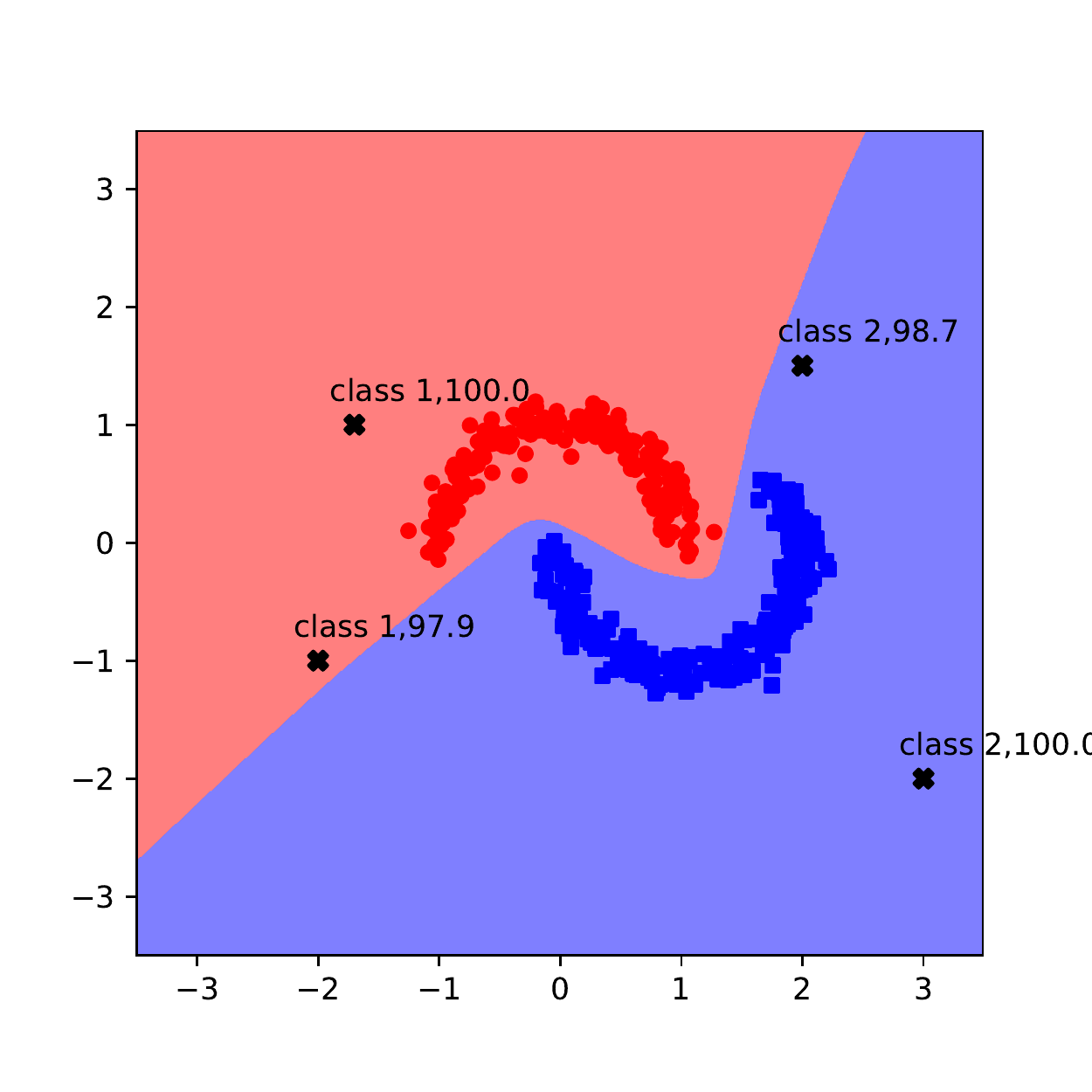}}~~
    \subfloat{\includegraphics[width=0.2\textwidth, trim= 1cm 0cm 0cm 1cm, clip=true]{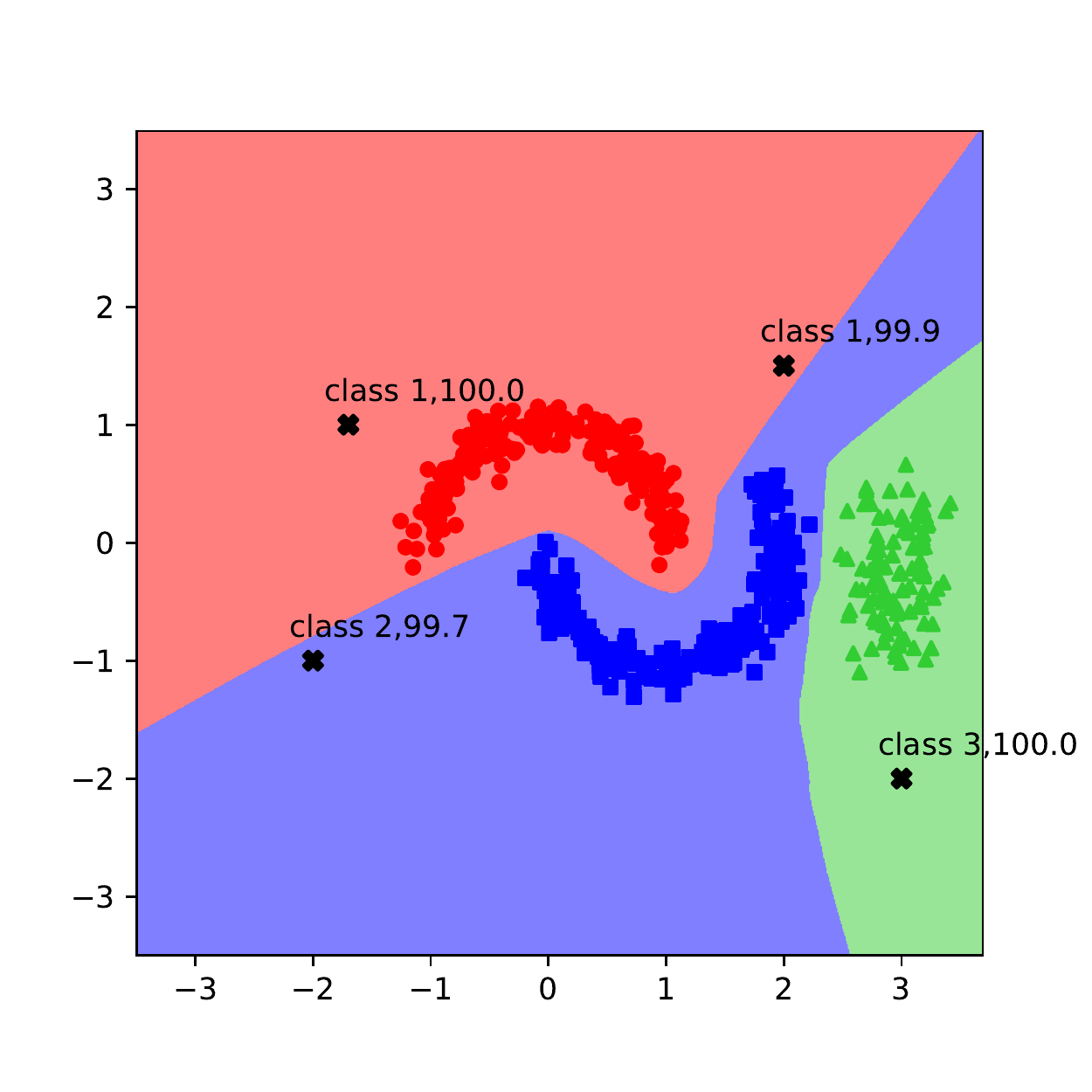}}~~
    \subfloat{\includegraphics[width=0.2\textwidth,trim= 1cm 0cm 0cm 1cm, clip=true]{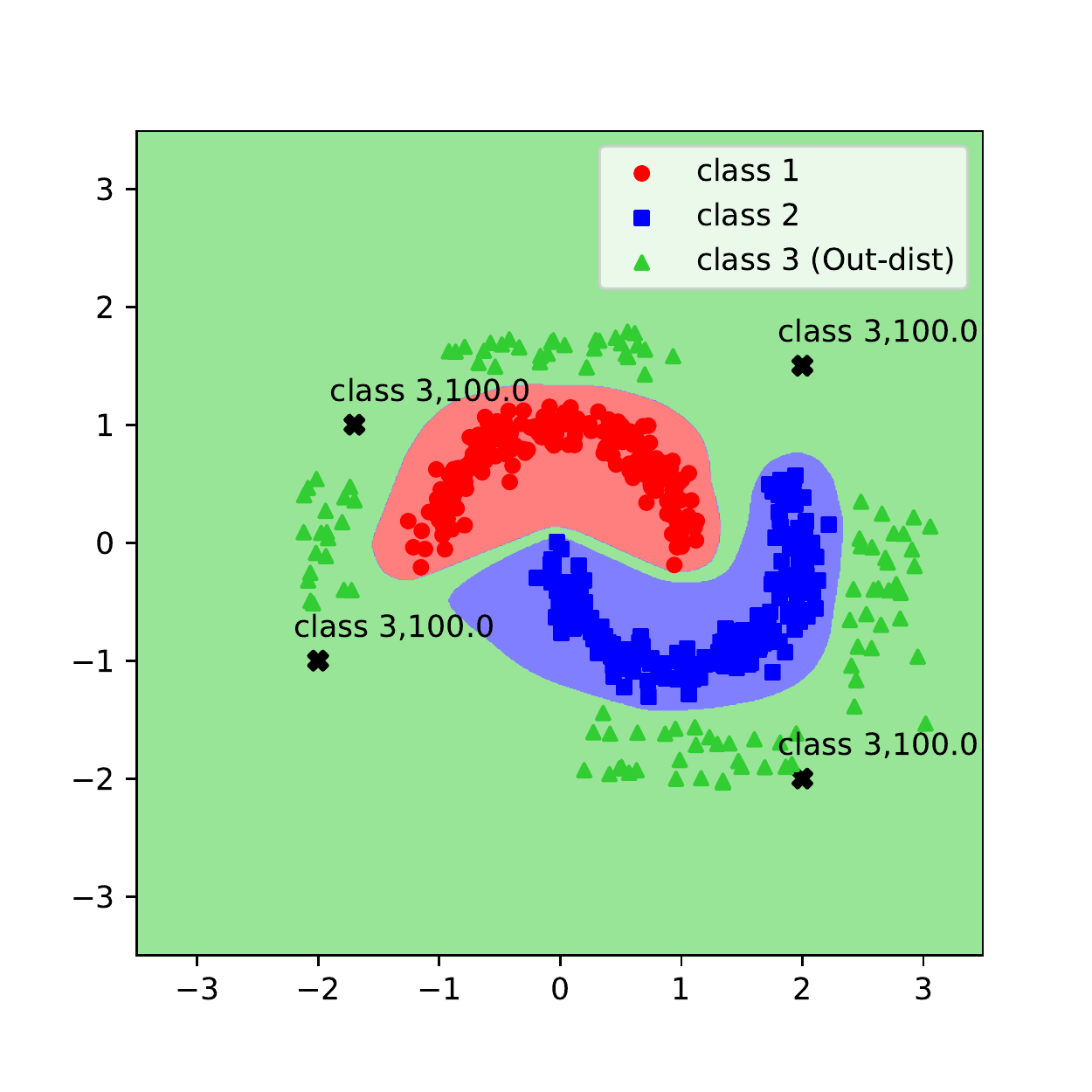}}
    \vspace{-1em}
    \caption{Illustration of properties of a partially-protective OOD set (middle) and a protective one (right) and their effect on training A-MLP for a two-moon classification dataset, compared to a (left) vanilla MLP trained on the same dataset. The black-cross samples are some test OOD samples and their predicted class and confidence scores by each (A)-MLP are also indicated. All MLPs are composed of three layers with ReLU activation function.}
 \label{toyexample}
 \end{figure*} 
 
\section{Characterizing a Proper OOD Set}
\label{proposedmethod}
 

Let us assume a hypothesis class $\mathcal{H'}$ (e.g. A-CNN) for a $K+1$ classification problem with $K$ classes associated for a given in-distribution task and the extra class (i.e., ($K+1$)-th class) reserved for identifying OOD samples. 
We also denote $\mathcal{S}_{I}=\{(\mathbf{x}_{I}^{i},\mathbf{y}_{I}^{i})\}_{i=1}^{N}$ as an in-distribution training set consisting of $N$ i.i.d.\ labeled samples drawn from data distribution $\mathcal{D}_I$, with true labels $\{\mathbf{y}_I^{i}\}_{i=1}^{N}\in \{1,\dots ,K\}$.
As the OOD training set, take $\mathcal{S}_O=\{(\mathbf{x}_{O}^{j})\}_{j=1}^{M}$ involving $M$ i.i.d.\ samples drawn from a data distribution $\mathcal{D}_O$, which we label as ($K+1$)-th class.

The loss of a hypothesis $h'\in\mathcal{H'}$ for a given in-distribution sample can be defined as $\ell (h'(\mathbf{x}_I^{i}),\mathbf{y}_I^{i})=\mathbb{I}(h'(\mathbf{x}_{I}^{i})\neq \mathbf{y}_{I}^{i})$ and its loss for an OOD sample is $\ell (h'(\mathbf{x}_{O}^{j}),K+1) = \mathbb{I}(h'(\mathbf{x}_{O}^{j})\neq K+1)$\footnote{{Indicator function $\mathbb{I}(p)$ returns $1$ if condition $p$ is true, and $0$ otherwise.}}. The \emph{true} loss of an augmented classifier $h'\in\mathcal{H'}$ can be evaluated on the underlying data distributions $\mathcal{D}_{I}$ and $\mathcal{D}_{O}$ as:
\begin{align}
L_{\mathcal{D}_{I}}(h') & =\mathbb{E}_{(\mathbf{x}_{I},\mathbf{y}_{I})\sim \mathcal{D}_I}\,\ell(h'(\mathbf{x}_I),\mathbf{y}_I),\\
L_{\mathcal{D}_{O}}(h') & =\mathbb{E}_{\mathbf{x}_{O}\sim \mathcal{D}_O}\,\ell(h'(\mathbf{x}_{O}),K+1 ).
\end{align}
The corresponding  \emph{empirical} loss is computed on training set $\mathcal{S}_{I}$ and $\mathcal{S}_{O}$:
\begin{align}
L_{\mathcal{S}_{I}}(h') & = \frac{1}{N}\sum_{i=1}^{N}\ell(h'(\mathbf{x}_{I}^{i}),\mathbf{y}_{I}^{i}),\\
L_{\mathcal{S}_O}(h')   & =\frac{1}{M}\sum_{j=1}^{M}\ell (h'(\mathbf{x}_{O}^{j}),K+1).
\end{align}

Before presenting our definition, we remark that there is a set of $B$ ``out'' data distributions $\mathcal{D}_{O}^{b}, b=\{1,
\ldots,B\}$ with their respective OOD training set $\mathcal{S}_{O}^{b}\sim\mathcal{D}_{O}^{b}$. Theoretically speaking, $B$ can be infinitely large. Moreover, we assume generalization error of vanilla classifier (denoted by $h\in \mathcal{H}$), for the original $K$ classification task, trained on $\mathcal{S}_{I}$ is less than a small $\epsilon$ value:
$|L_{\mathcal{S}_{I}}(h)-L_{\mathcal{D}_I}(h)| \le\epsilon$.

\begin{mydef}:
For a given OOD training set $\mathcal{S}_O^{b}\sim \mathcal{D}_{O}^{b}$ and in-distribution training set $\mathcal{S}_{I}$ w.r.t.\ hypothesis class $\mathcal{H'}$, $\mathcal{D}_{I}$ and $B$ ``out'' data distributions, we define two kinds of gaps for the augmented classifier $h'_{b}\in \mathcal{H'}$ trained on ${\mathcal{S}_{I} \cup \mathcal{S}_{O}^{b}}$, i.e. $\min_{h'_{b}} L_{\mathcal{S}_I}+ L_{\mathcal{S}_{O}^{b}}$ :

\begin{align}
\bm{\mathcal{L}}_{\mathcal{S}_I} & = |L_{\mathcal{S}_{I}}(h'_{b})-L_{\mathcal{D}_I}(h'_{b})|,\\
\bm{\mathcal{L}}_{\mathcal{S}_{O}^{b}}  & = \sup_{\mathcal{D}_{O}\in \{\mathcal{D}_{O}^{1}, \dots \mathcal{D}_{O}^{B}\}}|L_{\mathcal{S}^{b}_{O}}(h'_{b}) - L_{\mathcal{D}_{O}}(h'_{b})|.\label{eq_out2}
\end{align}
\end{mydef}



The first term $\bm{\mathcal{L}}_{\mathcal{S}_I}$ represents the gap between empirical loss of classifier $h'_{b}\in \mathcal{H'}$ on in-distribution training set $\mathcal{S}_I$ and its true loss on $\mathcal{D}_I$ while the second term $\bm{\mathcal{L}}_{\mathcal{S}^{b}_{O}}$ concerns the largest (worst) gap between empirical loss of $h'_{b}$ on OOD training set $\mathcal{S}^{b}_{O}$ and its true loss on ``out'' data distributions. By restricting $B$ to a manageable (finite) large number, we re-define $\bm{\mathcal{L}}_{\mathcal{S}_{O}^{b}}$ by upper-bounding Eq.~\ref{eq_out2}, i.e. sum of gaps on $B$ finite ``out'' data distributions:
\begin{equation}
    \bm{\mathcal{L}}_{\mathcal{S}_{O}^{b}}= \sum_{\mathcal{D}_{O}\in \{\mathcal{D}_{O}^{1}, \dots \mathcal{D}_{O}^{B}\}} |L_{\mathcal{S}^{b}_{O}}(h'_{b}) - L_{\mathcal{D}_{O}}(h'_{b})|.
\end{equation}

As true data distributions are unknown, the aforementioned equations can be empirically computed using validation sets. Then, a proper OOD set is the OOD set that training a A-CNN on it should produce the lowest accumulation of generalization errors of  both in-distribution task and (un)seen OOD sets:
\begin{equation}
    \mathcal{S}^{b^*}_{O}=\argmin_{\mathcal{S}^{b}_O\in\{\mathcal{S}_{O}^{1},\dots \mathcal{S}_{O}^{B}\}} \,\bm{\mathcal{L}}_{\mathcal{S}_{I}} +\lambda\bm{\mathcal{L}}_{\mathcal{S}_{O}^{b}},
    \label{optimal-out1}
\end{equation}
 where $\lambda>0$ is a balancing hyper-parameter. Directly using Eq.~\ref{optimal-out1} to find a proper OOD set is computationally inefficient as it involves training $B$ individual augmented classifiers, i.e. train each $h'_{b}$ on a pair of $\mathcal{S}_{I}\cup \mathcal{S}_{O}^{b},b\in\{1,\dots,B\}$. Particularly for the case of CNNs, this incurs a huge computational overhead. To overcome this computational burden, we conjecture that a protective OOD set can also provide a well-generalized A-CNN on both in- and unseen OOD sets (an intuitive illustration is given in Sec.~\ref{intuition}). Thus, instead of directly optimizing Eq~\ref{optimal-out1}, we develop some cost-effective metrics to assess the protectiveness level of OOD sets for identifying the most protective one.

\subsection{Protective OOD set: An Illustrative Example}\label{intuition}

To give a high-level intuitive explanation of our proposed metrics for recognizing a protective OOD set, we use an example based on the two-moon dataset (as in-distribution task), where each moon is considered as a sub-manifold. Fig.~\ref{toyexample}(a) exhibits the challenge of OOD samples for a vanilla MLP, which is trained on only in-distribution samples. As can be seen, this vanilla MLP \textit{confidently} classifies OOD samples (indicated with black-crosses) as either ``class 1'' or ``class 2'' albeit they clearly belong to \emph{none} of the in-distribution manifolds. In Fig.~\ref{toyexample}(b) we demonstrate a \textit{partially-protective} OOD set whose samples are almost collapsed and only partially cover one of the sub-manifolds (i.e., the manifold with blue squares). An augmented MLP (A-MLP) trained on the two-moon dataset along with this OOD set leads to a classifier with a limited OOD detection performance (lower generalization ability of detecting unseen OOD samples). More precisely, OOD samples, e.g., the \emph{unseen} black-cross samples, which are laying around \textit{uncovered} parts of the manifolds, are still confidently misclassified by the underlying A-MLP. Whereas, in Fig~\ref{toyexample}(c) a proper \textit{protective} OOD set, whose samples better cover the in-distribution's sub-manifolds (two-moon), is shown. As can be seen, training an A-MLP on such a protective OOD set (along with in-distribution samples) leads to classifying \emph{unseen} black cross OOD samples as class 3 (i.e., the extra class) as well as classifying automatically the regions out of the manifolds as class 3. This results in an A-MLP with high detection performance on unseen OOD sets (i.e., making the gap in Eq.~\ref{eq_out2} small).
Therefore, the design of our metrics is driven according to this intuition that a proper OOD set should be more protective of (i.e., closely covers) all in-distribution sub-manifolds in the feature space. A similar intuition has been previously exploited by some researchers, e.g. ~\cite{lee2017training,yu2017open}, with the aim of generating synthetic OOD samples.

\subsection{Proposed Metrics}
As previously done~\cite{feinman2017detecting,huang2006large}, we consider the penultimate layer of a vanilla CNN as a function that transfers samples from high-dimensional input space into a low-dimensional feature space, placing them on data (sub-)manifold(s)~\cite{bengio2013deep,brahma2015deep}. Furthermore, we assume that for a standard multi-classification problem (with $K$ classes), each class has its own sub-manifold in the feature space where its associated in-distribution samples lie. In the following, we propose our metrics to assess which of the available OOD sets has a better and closer coverage of the sub-manifolds.

\subsubsection{Softmax-based Entropy}
Our first metric aims at determining whether the samples of a given OOD set are distributed \emph{evenly} to all sub-manifolds (of a given in-distribution task) such that they have the equal chance of being covered by these OOD samples.
For example, an OOD set, whose samples are misclassified by a given vanilla CNN into \emph{only} a few of in-distribution classes (manifolds) instead of all of them, is deemed as a non-protective OOD set. This is because the sub-manifolds with no or only a few OOD samples being misclassified to them, are still uncovered (cf.\ Fig.~\ref{toyexample}(b)). Thus training A-CNN on such non-protective (or partially-protective) OOD set may lead to limited detection performance of unseen OOD sets. In contrast, the samples of a protective set are expected to be misclassified evenly to all the sub-manifolds, giving all of them an equal chance of being covered.

To quantitatively measure this incidence for a given OOD set w.r.t.\ an in-distribution set and a vanilla CNN trained on it, we introduce \emph{Softmax-based Entropy} (SE). First we define $p(c=k|{\mathcal{S}_O})$ as the conditional probability of $k$-th class given ${\mathcal{S}_O}$ as follows:
\begin{equation}
p(c=k|\mathcal{S}_{O})= \frac{1}{M}\sum_{j=1}^{M} \mathbb{I}_{k}\left( \argmax ( h(\mathbf{x}_O^{j}))=k\right),
\end{equation}
where ${h}$ is softmax output of the vanilla CNN trained on $\mathcal{S}_I$ and $\mathbb{I}_{k}(.)$ is an indicator function for $k$-th class. It returns $1$ if a given OOD sample $\mathbf{x}^{j}_{{O}}\in \mathcal{S}_O$ is (mis)classified as class $k$ by ${h}$, otherwise it returns $0$. 
Finally, SE is defined for an OOD set $\mathcal{S}_O$ as follows:
\begin{equation}
H(\mathcal{S}_{O})= -\sum_{k=1}^{K}p(c = k|\mathcal{S}_O)\log p(c = k|\mathcal{S}_O).
\end{equation}

$H(S_O)$ shall reflect how uniformly the samples of $\mathcal{S}_O$ are distributed to in-distribution sub-manifolds (i.e., corresponding to each in-distribution class). Note that the maximum value of SE is $\log K$ ($K$ is the number of classes) when all of the samples are uniformly distributed. Thus, the highest $H(\mathcal{S}_O)$ indicates that all the sub-manifolds have an equal number of OOD samples, whereas the smallest value of $H(\mathcal{S}_O)$ indicates some sub-manifolds (except a few of them) have a small number of (or no) OOD samples to cover them. Therefore, a protective OOD set should have a higher SE than that of non-protective ones.

\subsubsection{{Coverage Ratio}}

Although an OOD set with the high(est) SE confirms OOD samples are evenly distributed to all the sub-manifolds, using solely SE is not sufficient to assure the coverage of these sub-manifolds. Putting differently, an OOD set with the highest SE might still be collapsed and only partially cover some parts of the sub-manifolds. 

Inspired by covering number notion~\cite{shalev2014understanding}, we introduce our second metric, named coverage ratio (CR), in order to measure coverage of the sub-manifolds. Recall the sub-manifolds are approximated using a training in-distribution set in the feature space that is achieved by the penultimate layer of $h$. We denote $\mathbf{z}^{i}_{I}$ and $\mathbf{z}^{j}_{O}$ as the representations of $\mathbf{x}^{i}_{I}\in \mathcal{S}_{I}$ and $\mathbf{x}^{j}_{O} \in \mathcal{S}_{O}$ in the feature space, respectively.

To formally describe Coverage Ratio (CR), we form a rectangular weighted adjacency matrix $W \in \mathbb{R}^{N\times M}$ for a given pair $(\mathcal{S}_I, \mathcal{S}_O)$ with $N$ in-distribution and $M$ OOD samples, respectively. $W_{i,j}= \|\mathbf{z}^{i}_{I}-\mathbf{z}^{j}_{{O}}\|_{2} $ is the distance ($l_2$-norm) between in-distribution sample $\mathbf{z}^{i}_{I}$ and OOD sample $\mathbf{z}^{j}_{{O}}$ in the feature space. The distance between a pair of $(\mathbf{z}_I^{i}, \mathbf{z}_{O}^{j})$ is computed only if $\mathbf{z}^{i}_{I}$ is among $k$-nearest in-distribution neighbors of $\mathbf{z}^{j}_{{O}}$, otherwise $W_{i,j}=0$:
\begin{equation}
W_{i,j} = 
\begin{cases}
\|\mathbf{z}^{i}_{I}-\mathbf{z}^{j}_{{O}}\|_{2} & \text{if }\mathbf{z}^{i}_{{I}}\in \text{k-NN}(\mathbf{z}^{j}_{{O}}, S_I) \\
0 & \text{otherwise}
\end{cases}.
\end{equation}

In other words, for each sample $\mathbf{z}^{j}_{O}$, we find its $k$-nearest neighbors from the in-distribution set $\mathcal{S}_I$ in \emph{the feature space}. Then, if the given $\mathbf{z}^{i}_I$ belongs to $k$-nearest in-distribution neighbors of $\mathbf{z}^{j}_{O}$, we set $W_{i,j}$ to their distance. From matrix $W$, we derive a binary adjacency matrix $A$ as follows; $A_{i,j} = \mathbb{I}(W_{i,j}>0)$. Now using matrix $A$, we define CR metric as follows:
\begin{equation}
R(\mathcal{S}_I,\mathcal{S}_O) = \frac{1}{N}\sum_{i=1}^{N}\mathbb{I}\left(\sum_{j=1}^{M} (A_{i,j})>0\right),
\end{equation}
where $\mathbb{I}(\sum_{j=1}^{M} (A_{i,j})>0)$ assesses whether the $i$-th in-distribution sample $\mathbf{z}^{i}_I$ covered at least one time by the OOD samples $\mathcal{S}_O$ in the feature space.
Basically, this metric measures how many in-distribution samples (percentage) are covered by at least one OOD samples from $\mathcal{S}_O$ in the feature space. Finally, \textbf{we estimate an OOD set w.r.t.\ a given in-distribution set is protective if it has both high SE and high CR.} 

It is important to note that SE and CR are complementary. As mentioned earlier, high SE of an OOD set without considering its CR is not sufficient for estimating the protective level of an OOD set. Similarly, from high CR alone without having high SE, an OOD set cannot be considered as a protective one. This is because, an OOD set with high CR but low SE is not distributed enough among all sub-manifolds and might cover a large portion of only a few sub-manifolds.

\subsubsection{Coverage Distance} 
Furthermore, to measure the distance between OOD set $\mathcal{S}_O$ and the in-distribution data sub-manifolds, the following distance metric, named Coverage Distance (CD), can be driven:
\begin{equation}
D(S_I, S_O) = \frac{\sum_{i,j}W_{ij}}{\sum_{i,j}A_{ij}}= \frac{1}{kM}\sum_{i,j}W_{ij}.
\label{coverage-distance}
\end{equation}
$D(\mathcal{S}_I,\mathcal{S}_O)$ shows average distance between OOD samples of $\mathcal{S}_O$ and their $k$ nearest neighbors from in-distribution set.

\noindent{\textbf{Selection of Protective OOD set}}: We remark that for final selection OOD set \emph{SE and CR play more important roles than CD since they indicate the degree of spread and protectiveness of OOD sets for the sub-manifolds while CD reveals the average distance of OOD set to the sub-manifolds.} Since our primary concern is about coverage of the sub-manifolds, we first assess SE and CR. In other words, the most protective OOD set should have the highest SE (preferably near to $\log K$) and the highest CR compared to those of the remaining OOD sets.  If one encounters some OOD sets that have (relatively) equal highest SE and CR, then their CDs can be considered for final selection --the OOD set with smaller CD can be selected.

%% file: Evaluation.tex
\section{Experimentation}
\label{sec:eval}

\begin{figure*}[h]
    \centering
    \resizebox{1\textwidth}{!}{
    
     \subfloat[SVHN: ISUN/C100]{\includegraphics[width=0.3\textwidth]{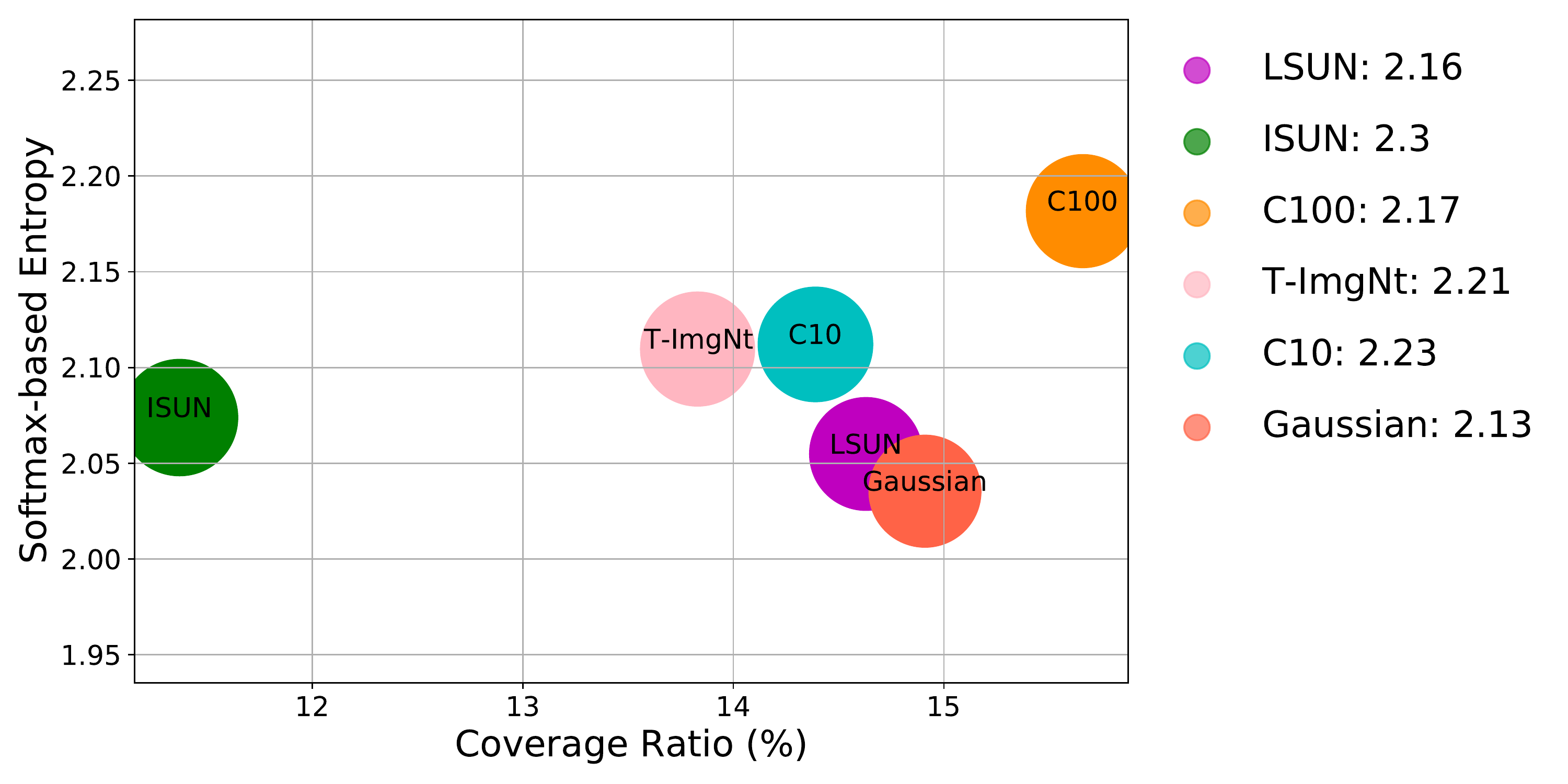}}~~\subfloat[CIFAR-10: SVHN/C100*]{ \includegraphics[width=0.3\textwidth]{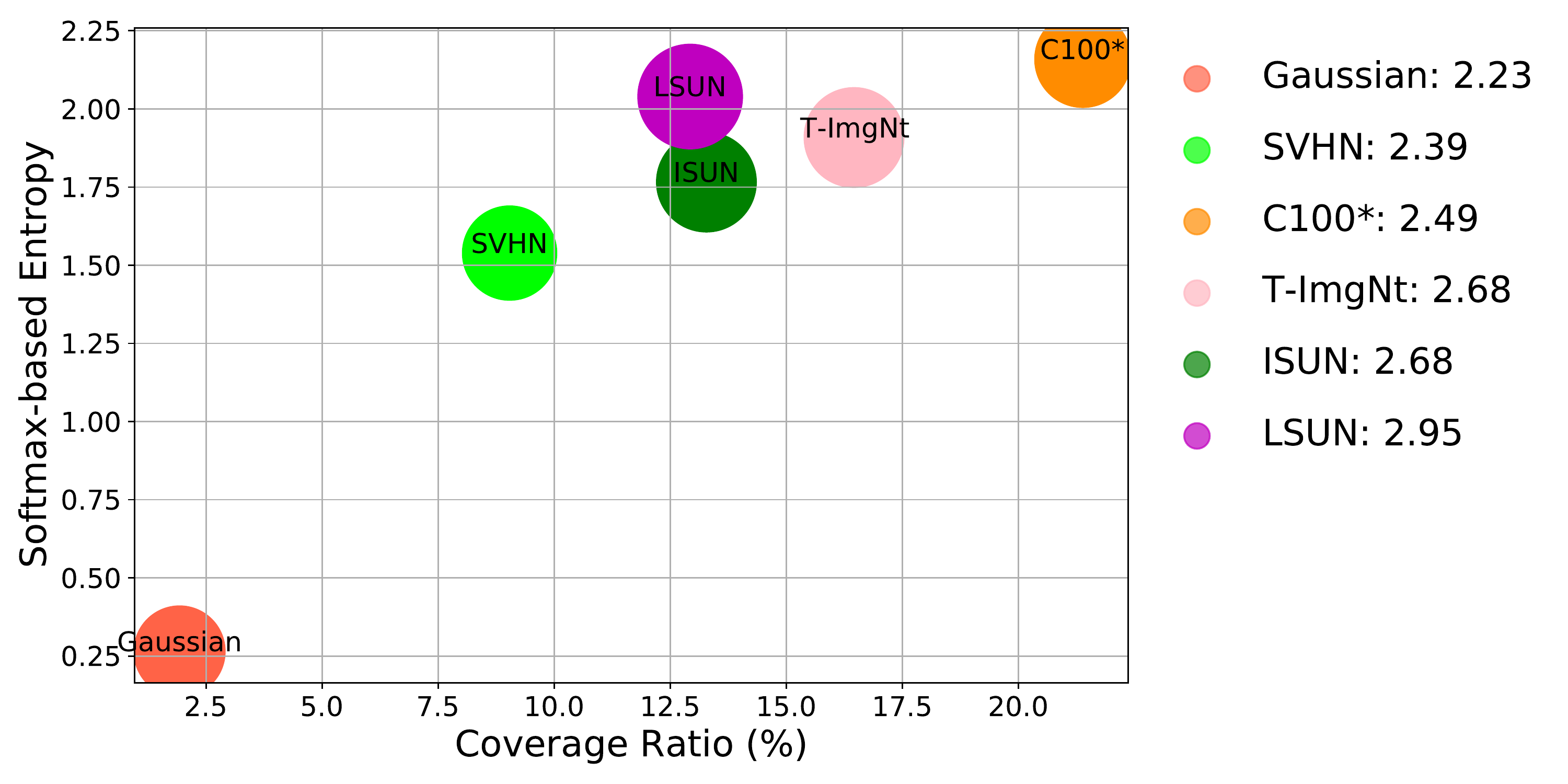}}~~\subfloat[Urban-Sound: (Command, TuT)/ ECS]{\includegraphics[width=0.3\textwidth]{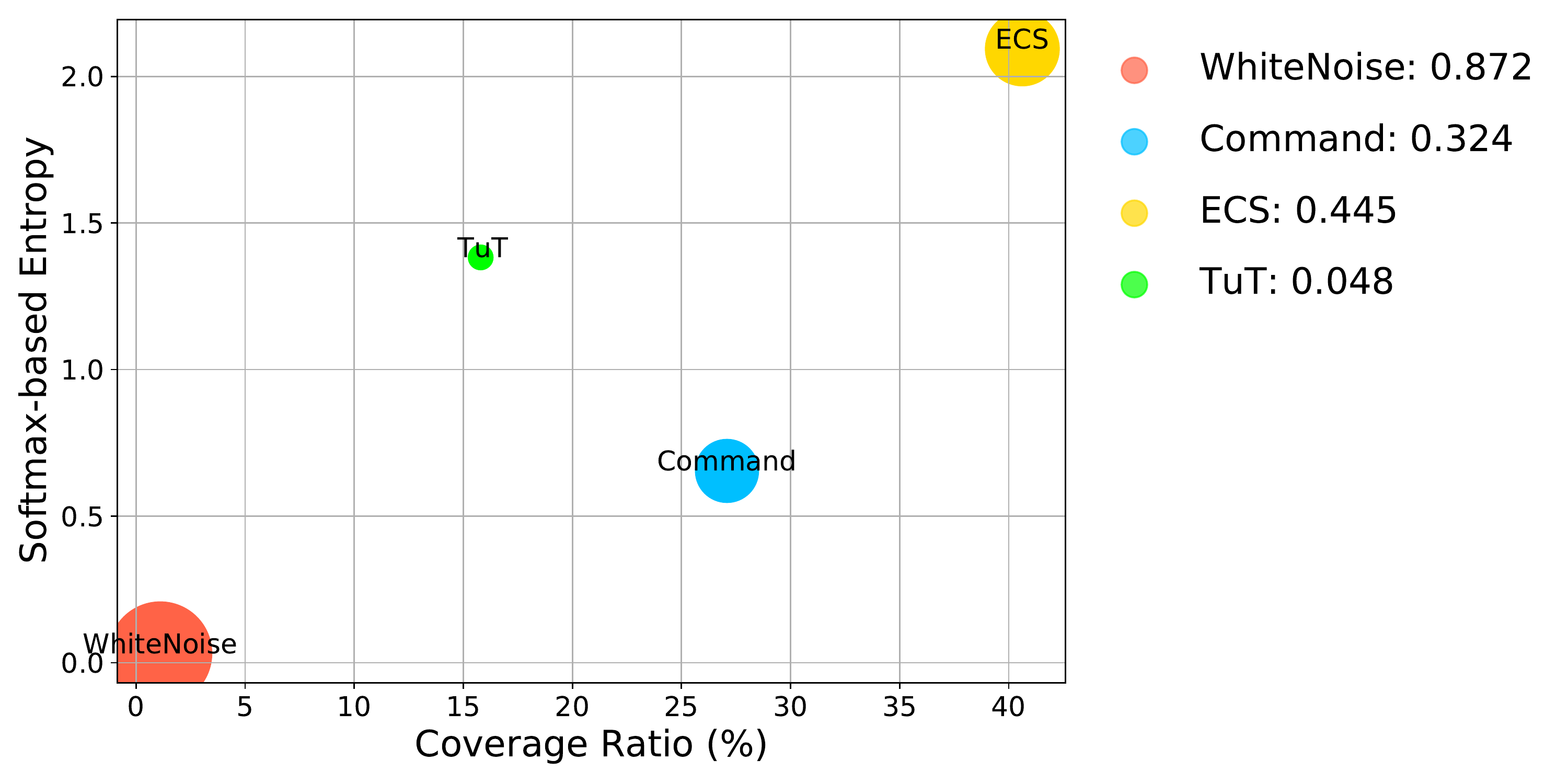}}}
    
    \caption{Differentiating OOD sets for SVHN, CIFAR-10, and Urban-Sound for the purpose of selecting the most protective one using our proposed metrics. Each sub-figure shows a bubble chart with SE and CR as y-axis and x-axis, respectively. The size of bubbles is determined by CD, also shown in the legend of the sub-figures. The least/most protective OOD sets are indicated in caption of the sub-figures.}
    \label{metrics}
\end{figure*}
We conduct a series of experiments on several classification tasks including two image benchmarks, namely CIFAR-10 and SVHN, and one audio benchmark, namely Urban-Sound~\cite{salamon2014dataset}. In our experiments, we utilize VGG-16 and a CNN described in~\cite{salamon2017deep} for image and audio benchmarks, respectively.  

Like in~\cite{liang2017principled}, for each of these in-distribution task, various naturalistic OOD sets are considered; for image classification tasks, we consider LSUN, ISUN, CIFAR-100 and TinyImageNet as OOD sets and Gaussian noise as a synthetic OOD set. 
For {audio classification task} with 10 classes, i.e., Urban-Sound, OOD sets considered are TuT~\cite{Mesaros2016_EUSIPCO}, Google Command~\cite{warden2018speech} and ECS (Environmental Sound Classification)~\cite{piczak2015dataset}, as well as white-noise sound as a synthetic OOD set. Note the classes of an OOD set, which are semantically or exactly overlapping with those of the given in-distribution set, are discarded.

In our experiments, we consider two types of end-to-end approaches, i.e., an A-CNN and a confidence-calibrated vanilla CNN, for detecting OOD set.  The latter type (i.e., calibrated vanilla CNN), a CNN is said to be calibrated after being trained to predict OOD training samples with great uncertainty (i.e. uniform prediction) while confidently classifying correctly in-distribution training samples. To achieve this, instead of cross entropy loss, we apply the modified loss function used in~\cite{hendrycks2018deep,lee2017training}. As the calibrated CNNs are threshold-based models for performing OOD detection, likewise~\cite{hendrycks2018deep,liang2017principled,lee2018simple} to assess its performance, we report AUROC, FPR at 95\% TPR on test for in- and out-of-distribution sets.

As A-CNN has an explicit rejection option (i.e. threshold-free approach), we consider three criteria to assess its performance on the in-distribution test set: 1) {Accuracy rate (Acc.) $\uparrow$}, that is the rate of samples classified correctly as their true associated label ($\uparrow$ indicates the higher the better), 2) {Rejection rate (Rej.) $\downarrow$}, the rate of samples misclassified as dustbin ($\downarrow$ indicates the lower the better), 3) {Error rate (Err.) $\downarrow$}, therate of samples that are neither correctly classified nor rejected (Err. rate = 1- (Acc. rate + Rej. rate)). A-CNN performance on OOD sets is evaluated by I) {Rejection rate (Rej.) $\uparrow$}: percentage (rate) of OOD samples classified as dustbin, and II) {Error rate (Err.) $\downarrow$}: rate of OOD samples not classified as dustbin (Err.  = 1 - Rej.)
Note since A-CNNs is a threshold-free OOD detector there are no AUROC and AUPR values. Plus, A-CNN's OOD rejection rate and its in-distribution rejection rate are the same concepts as TNR (True Negative Rate) and FNR (False Negative Rate), respectively.

\subsection{Empirical Assessment of  Metrics}

 First, to obtain in-distribution sub-manifolds, if in-distribution training set has more than $10,000$ samples, we randomly select $10,000$ samples from it, otherwise, we use the whole in-distribution training set. Secondly, we pass the samples through the penultimate layer of a pre-trained vanilla CNN to map them into the feature space. The same procedure is done for the samples of an OOD set to transfer them to the feature space. To fairly compare OOD sets according to the metrics, we also randomly select equal number of OOD samples ($10,000$ samples ) from each OOD set. For OOD sets with various sizes, we take the minimum size for making equal size OOD sets by randomly selecting from them. To compute CR and CD metrics, we set our only hyper-parameter, i.e. the number of nearest neighbors, $k=4$ for all our experiments. Note that among our metrics, CR and CD are dependent on $k$ (the impact of $k$ on our metrics are presented later).

 \subsubsection{Differentiating OOD Sets by the Metrics }
Using the proposed metrics, we differentiate OOD sets for each in-distribution set to select the most protective OOD sets w.r.t.\ the given in-distribution. 
In Fig.~\ref{metrics}, we demonstrate the difference between OOD sets according to their SE, CR, and CD, in order to identify the most and least protective OOD set. The most and least protective naturalistic OOD sets identified by our metrics (particularly by SE and CR) are indicated in caption of sub-figures in Fig.~\ref{metrics}. For \emph{SVHN} task, for example, \emph{ISUN}, among naturalist OOD sets, and \emph{Gaussian noise}, as synthetic OOD set, are identified as the least protective sets. Note that despite the high CR of Gaussian noise, its SE is far small, indicating it as a collapsed OOD set, which thus causes it to be identified as the least protective. The most protective OOD set for SVHN is CIFAR-100 (i.e., C100) with the highest SE and CR. For Urban-sound, ECS is the most protective, while Command and TuT datasets can be regarded as non-protective OOD sets due to their significantly low SE with respect to the upper bound of SE ($\log 10$). 
 
\begin{figure}[h]
\vspace{-2em}
     \centering
     \resizebox{0.5\textwidth}{!}{
     \subfloat{\includegraphics[width=0.25\textwidth]{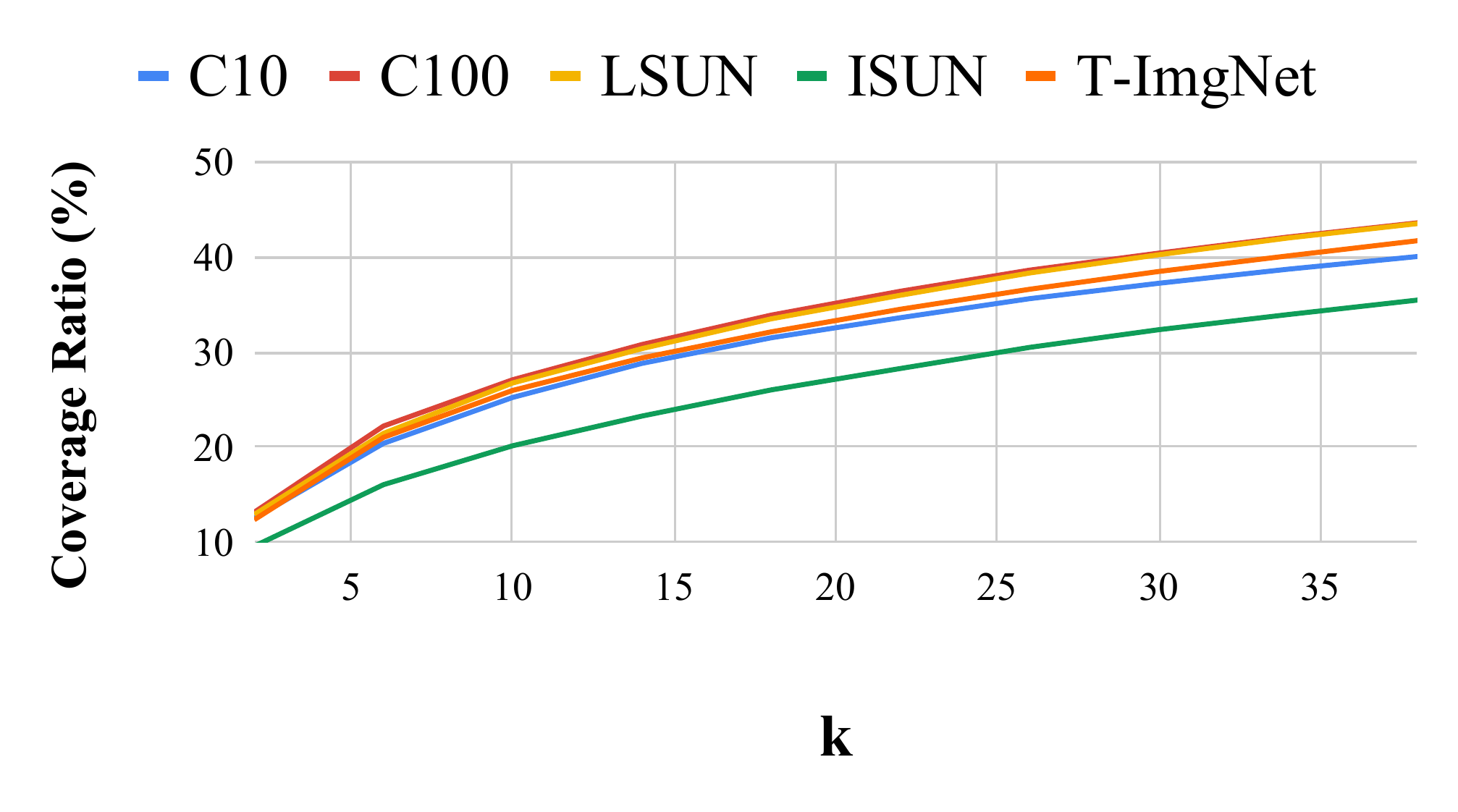}}~~\subfloat{\includegraphics[width=0.25\textwidth]{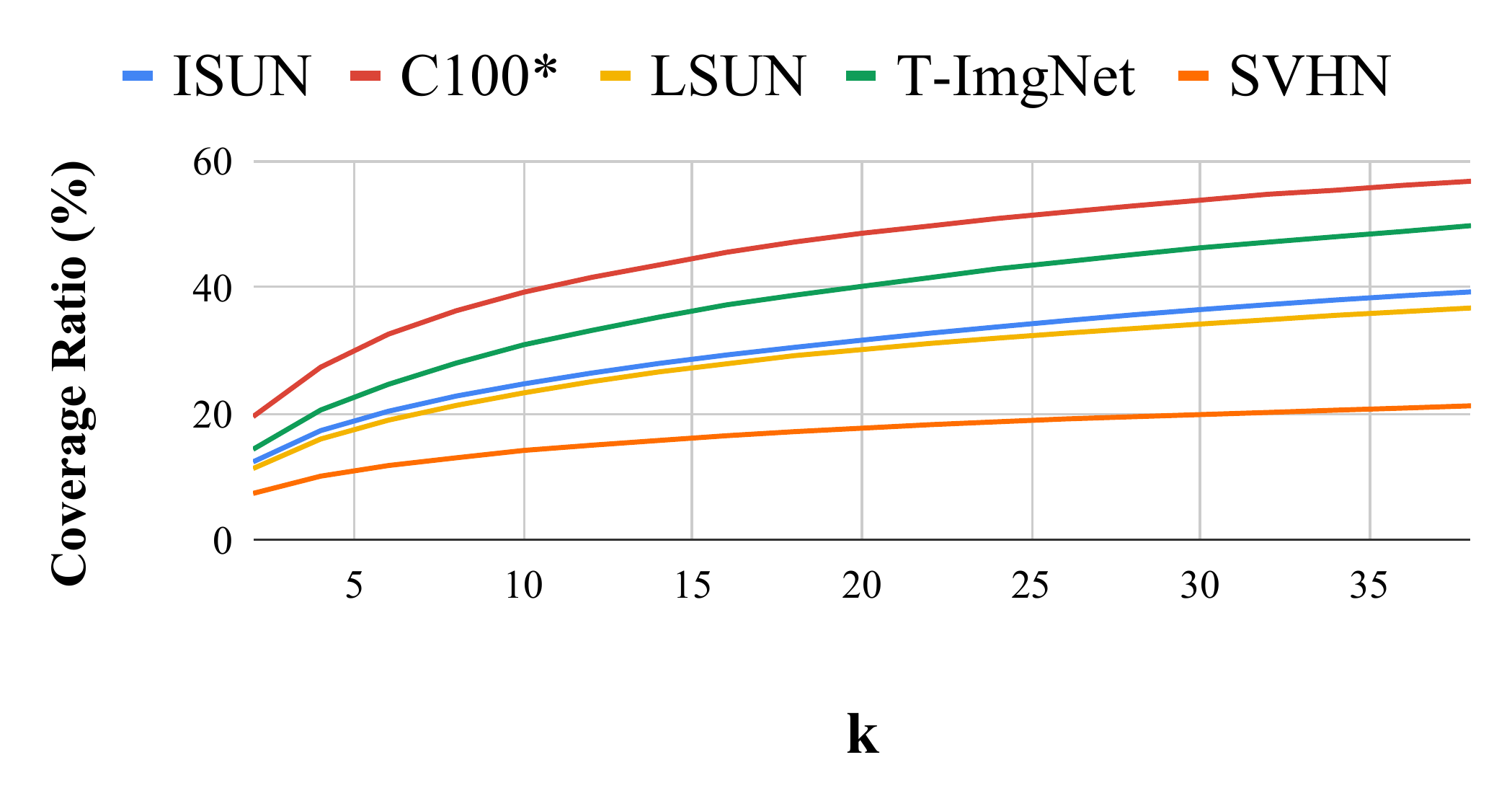}}
     }
     \vspace{-2em}
     \caption{ The effect of $k$ (number of nearest neighbors) on CR of OOD sets for SVHN (left) and CIFAR-10 (right).}
     \label{Sens-k}
\end{figure}

To assess the sensitivity of our metrics to the choice of $k$, we show the CR of OOD sets for varying $k$'s values in Fig~\ref{Sens-k}. In our experiments, we observe that the relative ranking of OOD sets according to their CR and CD is consistent with various values of $k$. Thus, CR and CD are not sensitive to the choice of $k$.
\begin{figure*}
    \centering
    
     \subfloat[SVHN]{ \includegraphics[width=0.3\textwidth]{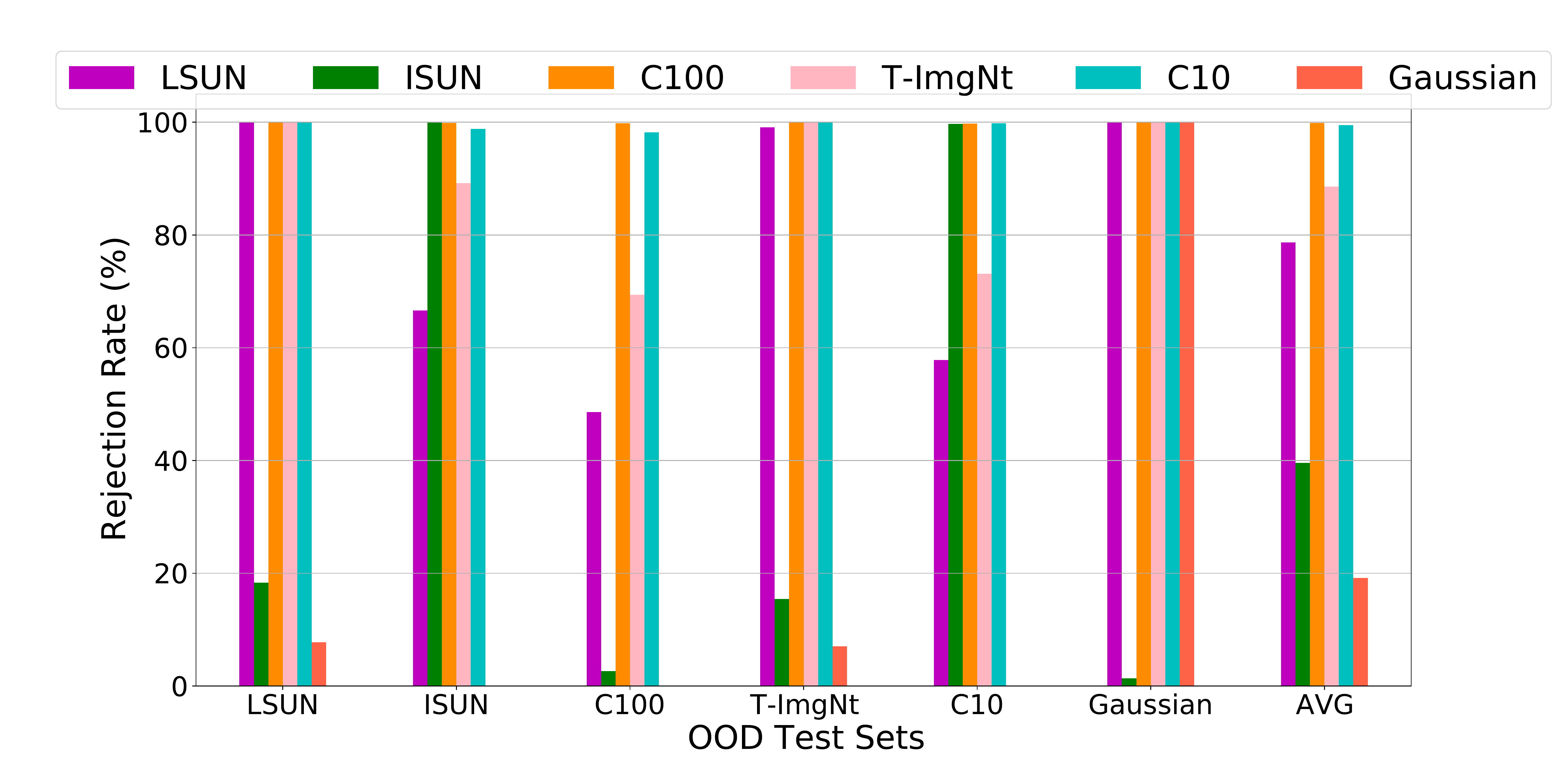}}~~\subfloat[Cifar10]{\includegraphics[width=0.3\textwidth]{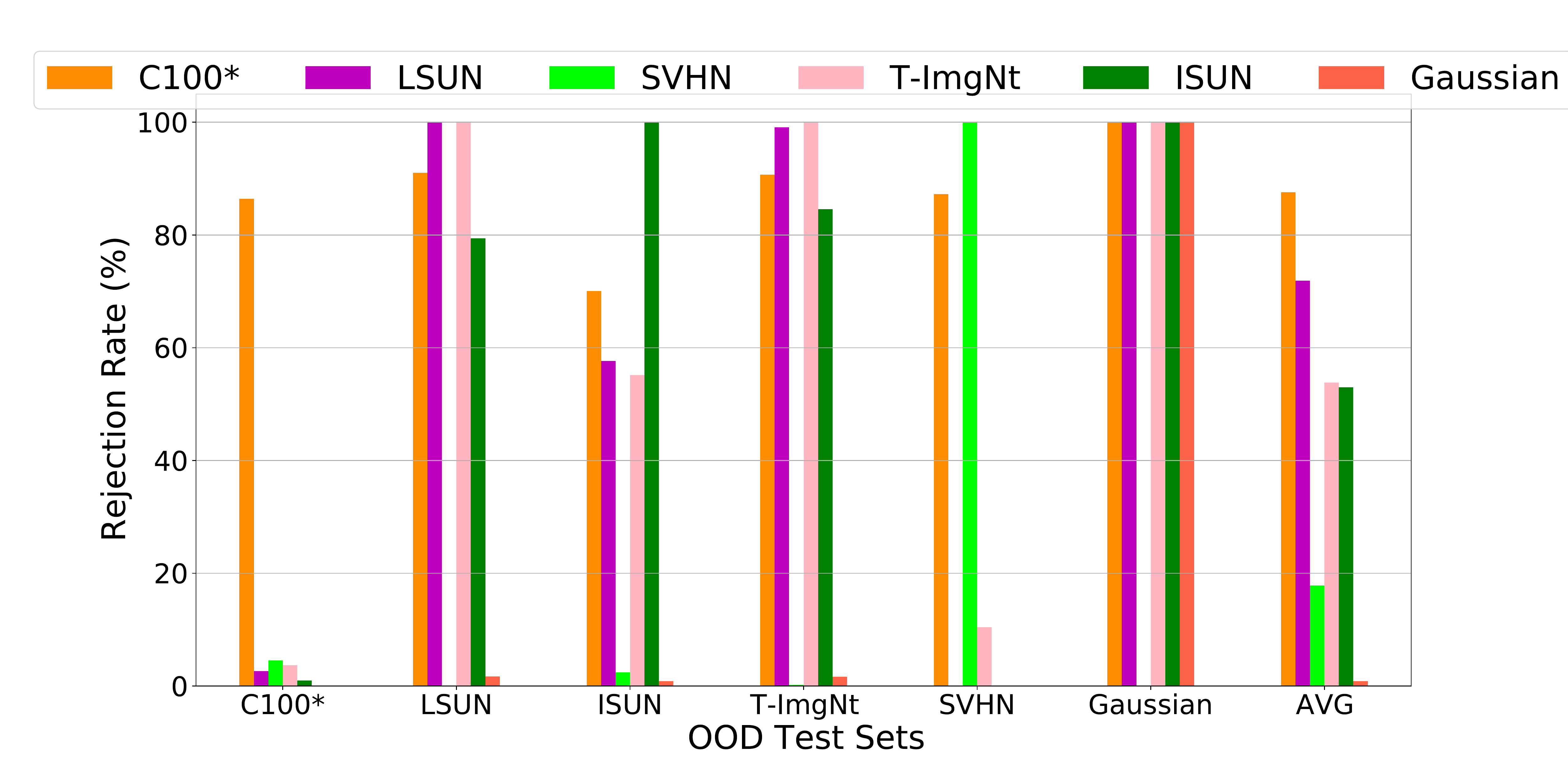}}
    ~~\subfloat[Urban-Sound]{\includegraphics[width=0.3\textwidth]{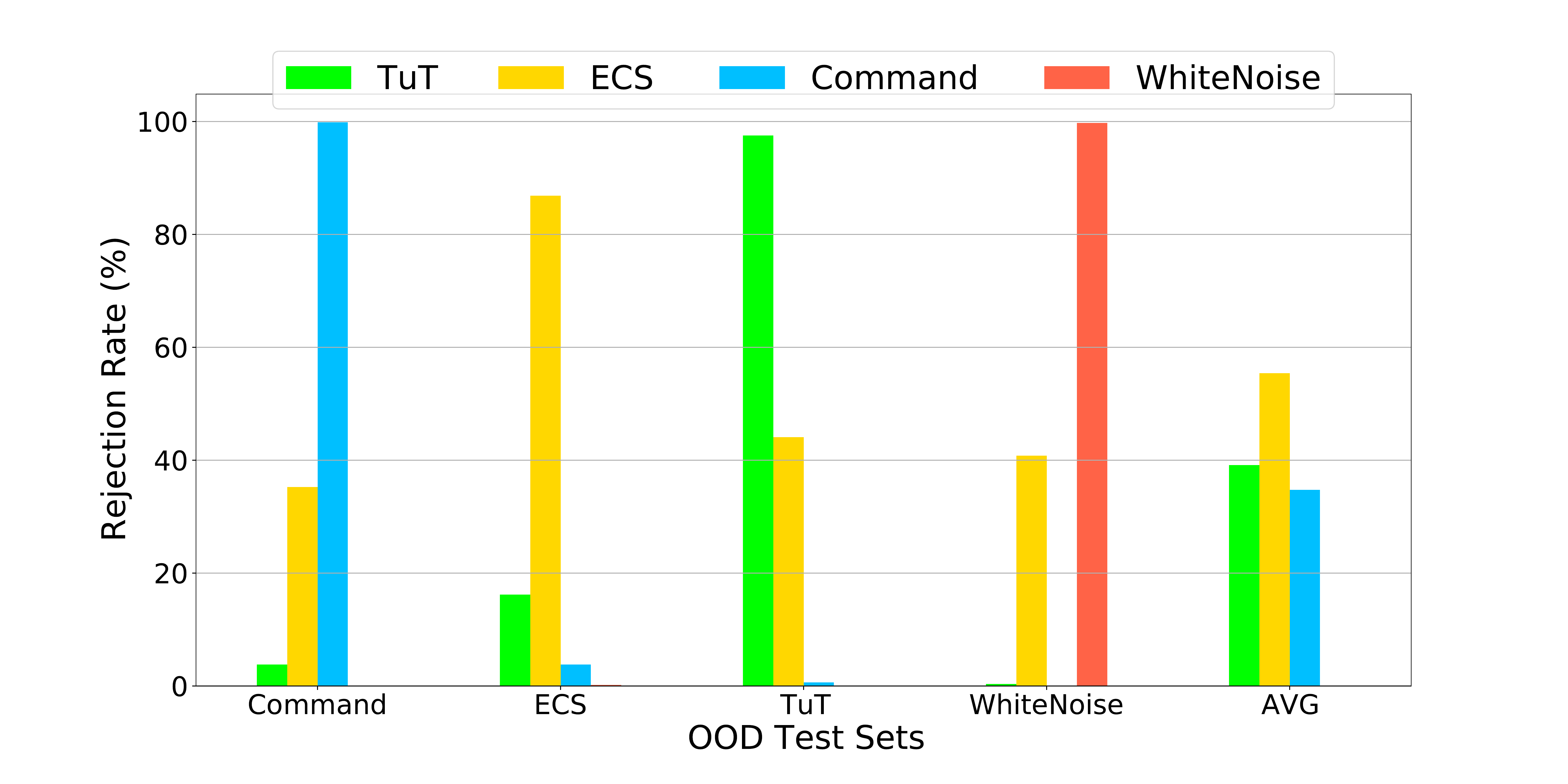}}
    \caption{ Rejection rates of A-CNNs on different OOD sets (x-axis), where each A-CNN is trained on a single OOD set, e.g. for SVHN in Figure (a) A-VGG(C100) means an Augmented VGG trained on SVHN as in distribution and CIFAR-100 as OOD set.  }
    \label{A-CNNBar}
\end{figure*}

{\textbf{A-CNN}}: As Eq.~\ref{optimal-out1} states, training an A-CNN on a proper OOD set should lead to a low average of \footnote{Instead of summation in Eq.~\ref{optimal-out1}, we take the average.} error rates on (un)seen OOD sets (or equivalently high average of OOD sample rejection rates). Therefore, for a given in-distribution task (e.g. SVHN), we train a separate A-CNN on each OOD set. Then, the error rates of these A-CNNs on all of the OOD sets are evaluated. Note a small error rate on a OOD set is equivalent to high detection rate.

In Fig~\ref{A-CNNBar}, A-CNNs trained on the most protective set (identified by our metrics) across various in-distribution tasks consistently outperform the A-CNNs trained on other OOD sets, particularly the A-CNN trained on the least protective one. For instance, A-CNN trained for CIFAR-10 with CIFAR-100* (the non-overlapped version of CIFAR-100) as the most protective OOD set has $85\%$ rejection rate on average while the least protective one, i.e. SVHN, delivers A-CNN with the lowest average of rejection rates ($21\%$) of OOD sets.

It is also interesting to note that even though one may expect Gaussian noise (i.e. white noise) to have well distributed samples, SE metric shows that its samples are actually not evenly distributed over all of the in-distributed sub-manifolds (having lowest SE) and sometimes (for CIFAR-10 and Urban-Sound in-distribution sets) they even have a small coverage rate. As a result, an A-CNN trained on Gaussian noise as an OOD set has the lowest average OOD rejection rate.

Consequently, the results show that all of the OOD sets are not equal for training well-generalized A-CNNs as they do not equally protect in-distribution sub-manifolds. \emph{Thus, we highlight that protectiveness can be an important factor for differentiating OOD sets to ultimately select the most proper one.} Moreover, to select an such OOD set, we remark that our metrics are computationally inexpensive than explicitly optimizing Eq.~\ref{optimal-out1}, which is equivalent to searching exhaustively all A-CNNs, where each is trained on an OOD set.


In Table~\ref{indistribution}, \emph{in-distribution} generalization performance of two A-CNNs, one trained on the most protective OOD set (named A-CNN$^{\star}$) and another trained on the least protective one (named A-CNN$^{\ddagger}$), are compared with their standard (Vanilla) CNN. Although the accuracy rates of the A-CNNs$^{\star}$ drop slightly, their error rates (i.e., risks) are considerably smaller than their counterparts, i.e., vanilla CNNs. This is because the A-CNNs are able to reject some ``hard" in-distribution samples, instead of incorrectly classifying them (similar to ~\cite{geifman2019selectivenet}). Rejecting a sample rather than incorrectly classifying it is an important aspect, particularly for security and safety concerns.

\begin{table}
\centering
\resizebox{0.5\textwidth}{!}{

\begin{tabular}{cccc}
    
    & &In-distribution  & \multicolumn{1}{c}{OOD sets }\\
    
In-dist. task &Network & Acc ($\uparrow$) / Rej ($\downarrow$) / Err ($\downarrow$) & Avg OOD Rej. ($\uparrow$)   \\
\hline

\multirow{3}{*}{ SVHN } & Vanilla VGG & \textbf{95.53} / -- / 4.47 & -- \\
& A-VGG$^{\ddagger}$ (ISUN) & 95.11 / 0 / 4.89 & 47.23 \\
& A-VGG$^{\star}$ (C100) & 95.38 / 0.34 / \textbf{4.28} & \textbf{99.88}\\
\hline
\multirow{3}{*}{ CIFAR-10 } & Vanilla VGG & \textbf{88.04} / -- / 11.95 & -- \\
& A-VGG$^{\ddagger}$ (SVHN) & 87.75 / 0.03 / 12.22 & 21.41 \\
& A-VGG$^{\star}$ (C100*) & 85.37 / 5.65 / \textbf{8.97} & \textbf{85.10} \\
\hline

& Vanilla CNN & \textbf{67.27} / -- / 32.73 & -- \\
Urban-Sound & A-CNN$^{\ddagger}$ (Command) & 65.05 / 2.02 / 32.93& 26.07 \\

& A-CNN$^{\star}$ (ECS) & 63.13 / 12.02 / \textbf{24.85}& \textbf{55.40} \\
\end{tabular}}
\caption{{The influence of selected most and least protective OOD sets on inducing well-generalized A-CNNs with high OOD detection rates.}}
     \label{indistribution}
     
\end{table}

\noindent{\textbf{Confidence-Calibrated vanilla CNN:}} Instead of A-CNN, now we use calibrated CNN as the end-to-end model in order to show the different impacts of OOD sets on training well-performing calibrated CNNs.
 As it can be seen in Table~\ref{tab:calibratedCNN}, the most protective OOD set recognized by our metrics is leading to a calibrated CNN with a considerable lower average of FPR at 95\% TPR and highest AUROC. While the calibrated CNN training on the least protective one has the higher FPR and the lower AUROC\footnote{For brevity, we report the average of AUROCs and FPRs of \emph{unseen} OOD sets.}.
 As a result, we highlight that efficiently recognizing proper (i.e. protective) OOD sets among the enormous available ones is a key for training a well-performed \emph{end-to-end model} (either the underlying model is A-CNN or calibrated vanilla CNN). 
 \begin{table}
    \centering
    \begin{tabular}{ccc}
In-distribution & Seen OOD set& Unseen OOD sets   \\
&& Avg AUROC/ Avg FPR\\
\hline
\multirow{5}{*}{ SVHN }& $\ddagger$ISUN & 94.73/31.97 \\
& LSUN & 99.25/ 4.39  \\
& C10 & 99.75/0.41 \\

& T-ImgNt & 99.75/1.10  \\
& $\star$C100 & \textbf{99.86/0.07} \\
\hline
\multirow{5}{*}{ CIFAR-10 } &  $\ddagger$SVHN & 86.38 /75.04  \\

& ISUN & 86.20/77.03 \\
& LSUN & 93.31/ 38.59 \\
&T-ImgNt & \textbf{93.89}/34.44  \\
&$\star$C100* & 93.03/\textbf{26.13}  \\
\hline

\multirow{3}{*}{Urban-Sound}&$\ddagger$Command&59.15/63.06\\
&$\ddagger$TuT&45.40/85.08\\
&$\star$ECS&\textbf{71.41/60.67}\\

\end{tabular} 
    \caption{The effect of OOD set selection on the performance of calibrated CNNs, where each trained on an OOD set, then evaluated on unseen OOD sets. We report the average of AUROC and FPR of calibrated CNNs on unseen OOD sets and test in-distribution set.}
    \label{tab:calibratedCNN}
    \vspace{-1em}
\end{table}

\begin{figure*}[h]
    \centering
     \resizebox{1\textwidth}{!}{
    
    \subfloat[{Rejection rate of SVHN adversaries}]{\includegraphics[width=0.25\textwidth]{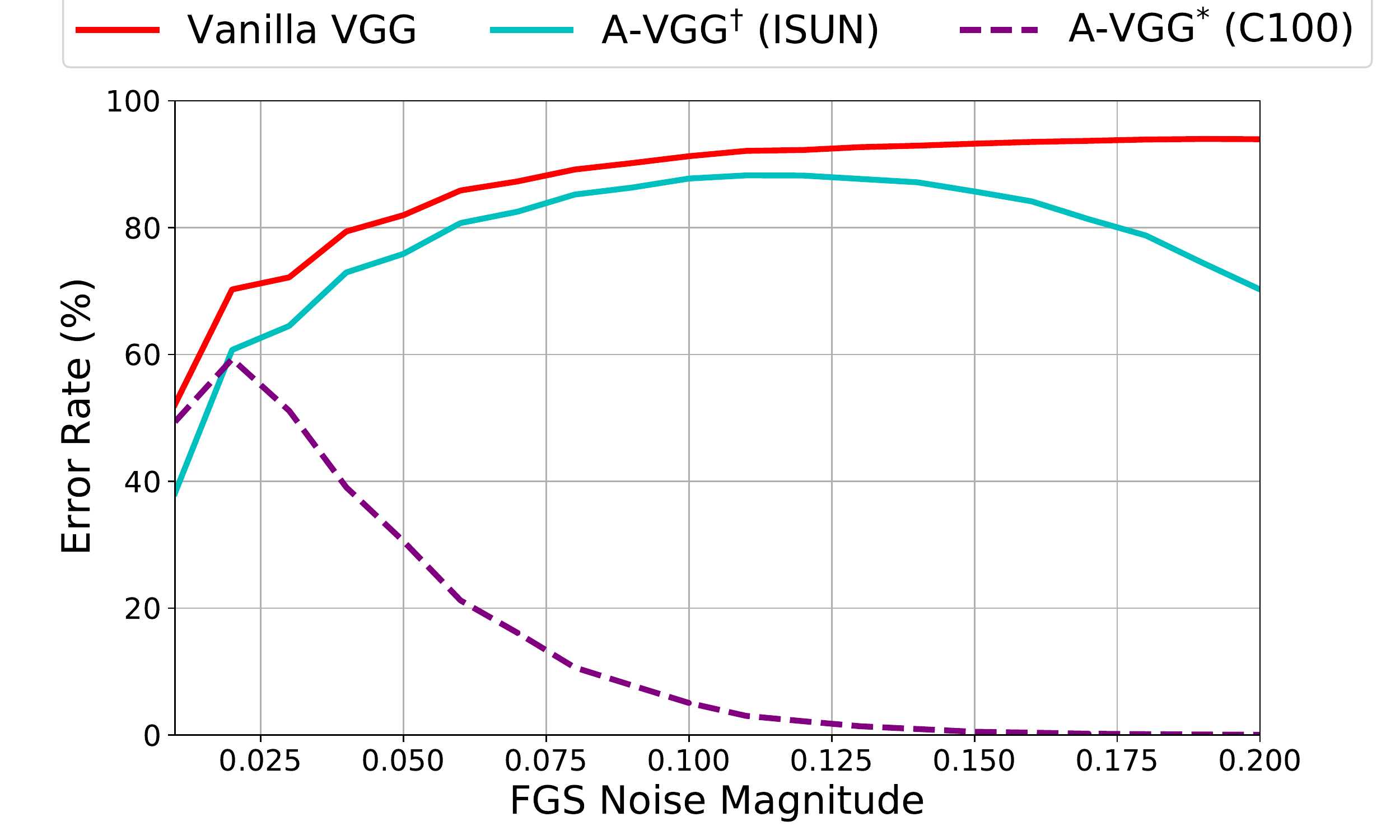}}~~\subfloat[{CD of SVHN adversaries}]{\includegraphics[width=0.25\textwidth]{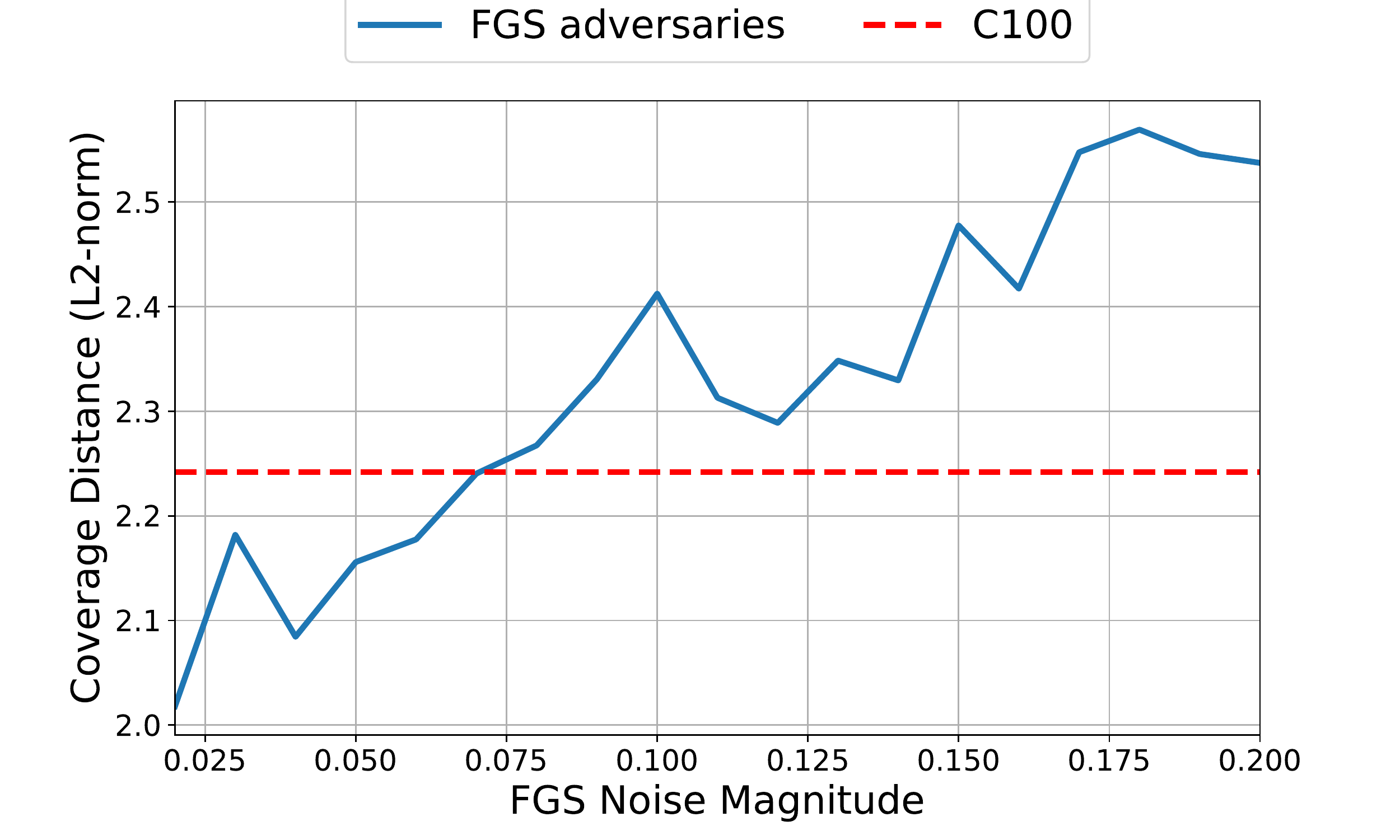}}~~
    \subfloat[{Rejection rate of CIFAR-10 adversaries}]{\includegraphics[width=0.25\textwidth]{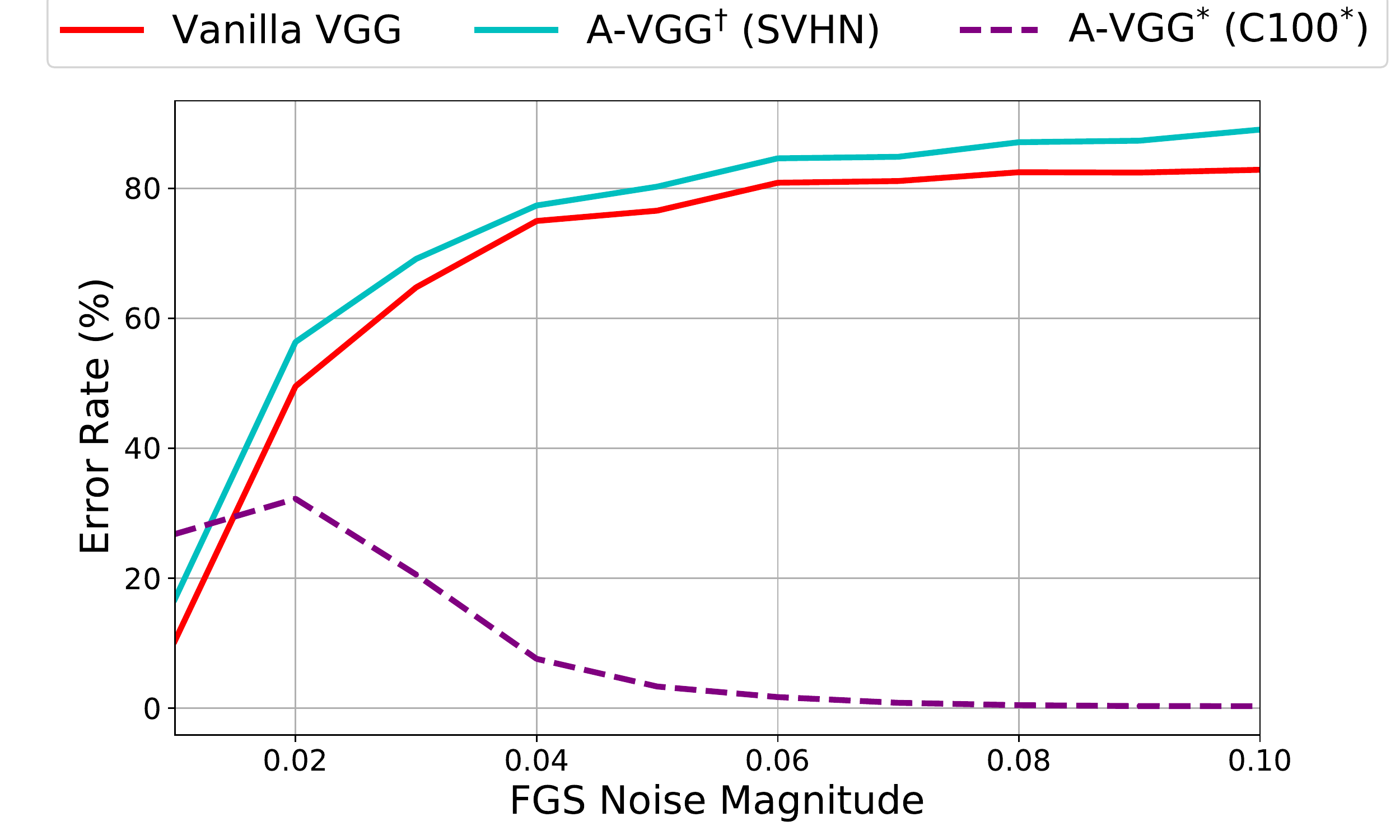}}~~\subfloat[{CD of CIFAR-10 adversaries}]{\includegraphics[width=0.25\textwidth]{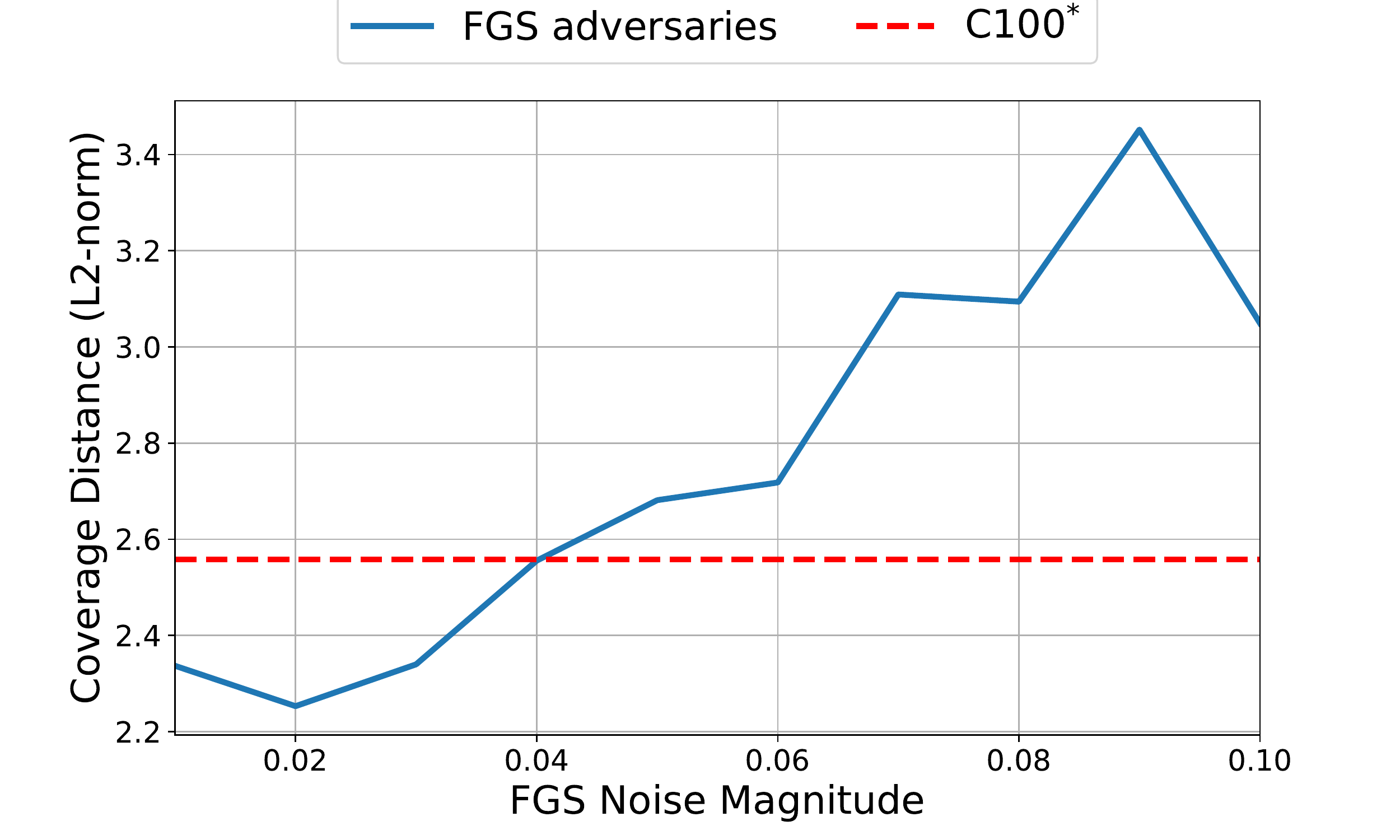}}
    }
    \\
     \resizebox{1\textwidth}{!}{
\begin{tabular}{cccccc}
     &\large{0.01}&\large{0.05}&\large{0.1}&\large{0.15}&\large{0.2}  \\
      \rotatebox{90}{\large{SVHN}} & \includegraphics[width=0.2\textwidth]{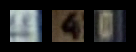}&
      \includegraphics[width=0.2\textwidth]{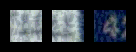}
      &
      \includegraphics[width=0.2\textwidth]{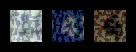}
      &
      \includegraphics[width=0.2\textwidth]{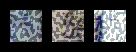}
      &\includegraphics[width=0.2\textwidth]{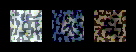}\\
      &&&&&\\
      &\large{0.02}&\large{0.04}&\large{0.06}&\large{0.08}&\large{0.1}  \\
      \rotatebox{90}{\large{C-10}}& \includegraphics[width=0.2\textwidth]{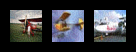}&
      \includegraphics[width=0.2\textwidth]{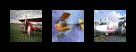}
      &
      \includegraphics[width=0.2\textwidth]{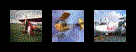}
      &
     \includegraphics[width=0.2\textwidth]{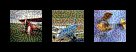}
      &\includegraphics[width=0.2\textwidth]{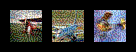}
 \end{tabular}}
  
    \caption{FGS adversaries with various noise magnitude (shown in the two last rows). Sub-figures (a,c) show error rates of vanilla CNN, A-CNN$^{\star}$, A-CNN${\ddagger}$ on FGS adversaries with varying noise for SVHN and CIFAR-10, respectively. Note Err rate = 1-(Acc rate +Rej rate). (b,d) Coverage Distance (CD) of FGS adversaries (their average distance to in-distribution sub-manifolds) for SVHN and CIFAR-10 respectively. The dotted red line  is the Coverage Distance of the most protective OOD set, which is used to train A-CNN$^{\star}$.}
 \label{adv-augmentedCNN}
 \end{figure*}

\subsection{Black-box Fast Gradient Sign (FGS) Adversaries as OOD samples}
FGS adversaries with high noise level can be regarded as synthetic OOD samples, where they most likely lie out of in-distribution sub-manifolds~\cite{goodfellow2014explaining,ma2018characterizing}. Even though such FGS adversaries contain perceptible noise, they can still fool vanilla CNNs easily~\cite{goodfellow2014explaining,tramer2017space}. To explore the capability of A-CNN$^{\star}$ in detecting such non-optimal adversaries, A-CNN$^{\star}$, A-CNN${\ddagger}$, and their vanilla counterparts are compared w.r.t. their error rates on FGS adversaries with a varying amount of noise. We generated $5,000$ black-box FGS adversaries (from training in-distribution set) using another pre-trained vanilla CNN (different from the one evaluated here). Some adversarial examples with various amounts of noise (i.e. $\alpha$) are displayed in Fig~\ref{adv-augmentedCNN}.

As evident from Fig~\ref{adv-augmentedCNN}, the error rates (i.e., 1-Acc) of vanilla CNNs increase as $\alpha$ becomes larger, showing the transferability of these black-box FGS adversaries. In contrast, the error rates (i.e., 1-(Acc+Rej)) of the A-CNNs$^{\star}$ approach zero as $\alpha$ increases since many of these FGS samples are rejected by A-CNNs$^{\star}$. On the contrary, the error rates of A-CNN${\ddagger}$ are almost as high as those of vanilla CNNs for FGS adversaries with different magnitudes of noise. Fig~\ref{adv-augmentedCNN} (b) and (d) can explain this phenomenon; larger $\alpha$ causes generated FGS adversaries to be further away from the sub-manifolds (i.e., larger CD). When FGS adversaries enter the protected regions by A-CNN$^{\star}$ (starting at the distance denoted by CD of the most protective OOD set, i.e., dotted red horizontal line), they are automatically rejected as OOD samples.

%% file: RelatedWork.tex
\section{Related Work}
\label{sec:relatedWork}

In~\cite{hendrycks2016baseline}, the authors have demonstrated that OOD samples can be discriminated from in-distribution ones by their predictive confidence scores provided by vanilla CNNs. As this baseline does not create a significant detection rate, many researchers~\cite{liang2017principled,lee2018simple,jiang2018trust,bendale2016towards,devries2018learning,lakshminarayanan2017simple} have attempted to process the confidence score for creating a larger gap between in-distribution and OOD samples. 

Other researchers have proposed to train end-to-end calibrated networks for making low confidence prediction on OOD sets while keeping in-distribution performance high. For example,~\cite{masana2018metric} have incorporated an OOD set to in-distribution set to train a modified Siamese network in order to keep in-distribution samples nearby while pushing OOD training samples away from in-distribution ones. Others~\cite{hendrycks2018deep,lee2017training,Vyas_2018_ECCV} have proposed to train a vanilla CNN on OOD set along with in-distribution set to force explicitly prediction of OOD samples with uncertainty while confidently and correctly classifying in-distribution samples. To train such end-to-end CNN-based models, one can leverage a natural OOD set likewise~\cite{bevandic2018discriminative,hendrycks2018deep,masana2018metric} or a set of synthetically-generated OOD samples~\cite{lee2017training,yu2017open,hein2019relu}. Apart from computational cost of generating such a set of synthetic samples, Hendrycks et al.\cite{hendrycks2018deep} have shown a calibrated CNN trained on a proper naturalistic OOD set can outperform that of trained on GAN-generated synthetic samples.

%% file: conclusion.tex
\section{Conclusion} \label{sec:conclusion}
Our main goal in this paper is to characterizing properties of OOD sets for recognizing a proper one for training an end-to-end A-CNN and calibrated vanilla CNN with high detection rate on unseen OOD sets while maintaining in-distribution generalization performance. To this end, we feature an OOD set as proper if it can cover all of the in-distribution's sub-manifolds in the feature space (i.e. protective OOD set). Then, we propose computationally efficient metrics as a tool for differentiating OOD sets for the purpose of selecting the most protective one. Finally, we empirically exhibit training end-to-end models on the most protective OOD set leads to remarkably higher detection rates of unseen OOD sets, in comparison with those models trained on the least protective OOD set. A Growing number of available OOD sets is a possible rich source for training well-performing end-to-end models to tackling OOD detection challenge, if the most proper OOD set (equivalently the most protective one) can be efficiently recognized.
\vspace{-.5em}
\subsection*{Acknowledgements}\vspace{-.5em}
This work was funded by NSERC-Canada, Mitacs, and Prompt-Qu\'ebec. We  thank Annette Schwerdtfeger for proofreading the paper.